\documentclass{article} % For LaTeX2e
\usepackage{booktabs}
\usepackage{iclr2026_conference,times}

\usepackage{amsmath,amsfonts,bm}

\def\eqref#1{equation~\ref{#1}}
\def\1{\bm{1}}

\DeclareMathAlphabet{\mathsfit}{\encodingdefault}{\sfdefault}{m}{sl}
\SetMathAlphabet{\mathsfit}{bold}{\encodingdefault}{\sfdefault}{bx}{n}

\usepackage{hyperref}
\usepackage{url}
\usepackage{pgfplots}
\usepackage{subcaption}
\usepackage{enumitem}
\pgfplotsset{compat=1.18}

\title{Bootstrapped Mixed Rewards for RL Post-Training: Injecting Canonical Action Order}

\author{Prakhar Gupta\thanks{Contact: \texttt{prakharg@umich.edu}} \\
University of Michigan \\
\And
Vaibhav Gupta
}

\usepackage{bbm}
\iclrfinalcopy
\begin{document}

\maketitle

\begin{abstract}
Post-training with reinforcement learning (RL) typically optimizes a single scalar objective and ignores structure in how solutions are produced. We ask whether a scalar hint toward a canonical solver ordering, used only during RL post-training, improves performance even when fine-tuned on randomized solution sequences. On Zebra puzzles, we fine-tune a Transformer on randomized solution orders, then post-train it with Group Relative Policy Optimization (GRPO) using two rewards: a sparse task reward that is 1 only when the puzzle is fully solved, and an ordering reward that increases when the model's emission order aligns with the canonical solver order. To compare signals cleanly, we combine them via fixed mixtures and use a simple bootstrapped scaling to equalize component magnitudes at initialization. Mixed rewards generally outperform task-only optimization, suggesting that coarse ordering signals can steer RL post-training toward canonical trajectories without modifying supervised data or architecture.\footnote{Code: \url{https://github.com/prakharg55/Zebra-ICLR-WM}}
\end{abstract}

\section{Introduction}

Post-training with reinforcement learning (RL) often optimizes simple scalar objectives (e.g., task success) but typically ignores structure in the environment, such as the order in which intermediate actions are taken---a setup widely used when post-training language models after fine-tuning \citep{NEURIPS2020_1f89885d,NEURIPS2022_b1efde53}.

We study Zebra puzzles (also known as Einstein's puzzles) with a GPT-2 style Transformer. We view each puzzle as a deterministic environment with latent state transitions; a policy that follows canonical solver-order trajectories can be viewed as behaving as if it maintains an internal model of valid next moves for each state. Prior work shows that Transformers trained on solver-ordered data encode the set of possible values in each cell that can be linearly decoded from hidden states, suggesting an implicit “reasoning engine” over puzzle dynamics \citep{NEURIPS2024_67b31ca1}. We ask whether a scalar hint toward solver ordering, used only in RL post-training, can improve performance even when fine-tuned on randomized solution orders.

The model is first fine-tuned on randomized solution orders. Next, we post-train with Group Relative Policy Optimization (GRPO) using two rewards: (i) a sparse solved reward (1 only if the model produces a fully correct solution), and (ii) an ordering reward that increases when the model's output order matches the solver order. To compare signals, we combine them via fixed mixtures and apply a bootstrapped reward scaling that equalizes component magnitudes at initialization to a target ratio.

Empirically, mixtures that include a non-zero ordering component generally outperform task-only optimization, with the best performance at a ${\text{solve}:\text{order}}$ weighting of $0.99:0.01$. Notably, the model never sees solver-order sequences during training; the ordering information is provided only as a scalar reward during RL, yet it still biases learning toward solver-order trajectories.

\paragraph{Contributions.} (1) A scalar reward hinting method that injects solver ordering into GRPO without modifying the fine-tuning data; (2) a bootstrapped scaling procedure that normalizes heterogeneous reward magnitudes, enabling controlled mixture studies; and (3) empirical evidence on Zebra puzzles that coarse ordering signals, when mixed with correctness, improve RL post-training accuracy.

\section{Data}

We use the Zebra puzzle dataset introduced in \citet{NEURIPS2024_67b31ca1}. Each example consists of a textual encoding of a Zebra puzzle followed by a sequence of actions that correspond to filling entries in an underlying entity-by-attribute grid. We filter the dataset to include only puzzles whose full solution has exactly $9$ actions, and each action in the solution sequence is encoded as a triplet $(\text{row}, \text{col}, \text{val})$. We consider two variants of the solution sequence:
\begin{enumerate}[label=(\roman*)]
\item a solver-order sequence, which records the exact order in which a canonical Zebra solver fills the solution cells, and
\item a random-order sequence, produced by shuffling the solver sequence uniformly at random.
\end{enumerate}

Following \citet{NEURIPS2024_67b31ca1}, the Zebra solver is a deterministic, human-like procedure that attempts to solve puzzles iteratively without backtracking. At each step, given the clues and partial grid, it tries to fill an entry whose value is uniquely determined under its fixed rule set, and repeats until solved. The solver-order sequence is the chronological list of triplets produced by this procedure.

The dataset design plays a crucial role in this work. Because the solver sequence reflects a valid step-by-step reasoning trajectory and the random sequence does not, we can measure and reward the degree to which the model follows solver ordering during generation. This choice is motivated by evidence that models trained on solver steps learn substantially more effectively than models trained on randomized steps \citep{NEURIPS2024_67b31ca1}.

\section{Training Setup}

We use a GPT-2 style Transformer as our base architecture. The model has 4 layers, 4 attention heads per layer, and a hidden size of 256. The model is trained from scratch rather than initialized from a pretrained checkpoint. Hyperparameters for our training can be found in Appendix~\ref{sec:appx_hyperparameters}.

\paragraph{Standard Fine-Tuning.} The model is first trained with a standard causal language modeling objective on the Zebra dataset. For each example, the puzzle is given as input, and the target output consists of the corresponding solution sequence. The loss is computed only over the solution tokens, with puzzle tokens masked out. We apply this procedure to fine-tune our baseline model on randomized solution orders.

\paragraph{GRPO Post-Training.} After fine-tuning, we further post-train the random-order model using Group Relative Policy Optimization (GRPO) \citep{shao2024deepseekmath}, a reinforcement learning algorithm that enables optimizing arbitrary reward functions over model rollouts. During GRPO training, rewards are computed over generated sequences and used to update the model via the GRPO loss. We evaluate several reward designs, including a solved reward, order reward, and their weighted combinations (described in Section ~\ref{sec:reward}).

\section{Reward Design} \label{sec:reward}

A key focus of this work is the design of reward functions that guide GRPO post-training. Our objective is to study whether scalar hints about the underlying solving order can improve solving accuracy without explicitly training on ordered sequences. We consider two primary reward functions and several fixed weighted combinations of the two. In all reward computations, we score only the completion prefix up to the ground-truth action length to keep indices comparable across rollouts.

\subsection{Solved Reward}

Let the ground-truth solution be the set of triplets $S^\star=\{(r,c,v)\}$. We define a sparse solved reward that is $1$ only when the model produces a fully correct solution and $0$ otherwise, capturing overall correctness but not the order of predicted cells. We parse the model completion into triplets, ignore any cell $(r,c)$ assigned multiple distinct values, and mark the rollout as solved only if every ground-truth cell is present with the correct value:
\begin{equation}
R_{\text{solve}} \;=\; \mathbbm{1}\!\left[\forall (r,c,v)\in S^\star:\ \hat{v}(r,c)=v \ \land\ \text{no conflicting values are assigned to }(r,c)\right].
\end{equation}

\subsection{Order Reward}

The order reward measures how closely the model's generation order matches the canonical solver order, independent of whether the predicted values are correct. Let $\pi^\star(r,c)$ denote the index of cell $(r,c)$ in the canonical solver solution, and let $\hat{\pi}(r,c)$ denote the index at which the model first emits cell $(r,c)$ in its completion. We only include cells that the model emits exactly once (to avoid ambiguity from repeated assignments). For eligible cells, we compute
\begin{equation}
r(r,c) \;=\; \frac{1}{1 + \lvert \pi^\star(r,c) - \hat{\pi}(r,c) \rvert}.
\end{equation}
We then average over eligible cells:
\begin{equation}
R_{\text{order}} \;=\; \frac{1}{\lvert \mathcal{C} \rvert}\sum_{(r,c)\in \mathcal{C}} r(r,c),
\end{equation}
where $\mathcal{C}$ is the set of ground-truth cells emitted exactly once. If $\mathcal{C}$ is empty, we set $R_{\text{order}}=0$.

This yields a reward-shaping signal \citep{10.5555/645528.657613} that guides toward solver-like rollouts without enforcing exact sequence reproduction.

\subsection{Combined Rewards and Bootstrapped Scaling}

We combine the two rewards via a fixed weighted sum:
\begin{equation}
R_{\text{total}} \;=\; \alpha \cdot R_{\text{solve}} \;+\; (1-\alpha) \cdot R_{\text{order}}, 
\qquad \alpha \in [0,1].
\end{equation}
Viewing $R_{\text{total}}$ as a weighted composition of objectives follows standard practice in multi-objective RL \citep{10.5555/2591248.2591251}. To avoid manual tuning when the raw magnitudes of $R_{\text{solve}}$ and $R_{\text{order}}$ differ, we employ a simple bootstrapped reward scaling prior to GRPO. We evaluate the frozen fine-tuned model on the validation split and compute mean rewards $\bar{R}_{\text{solve}}$ and $\bar{R}_{\text{order}}$. For a desired mixture $\alpha$, we set global scalars
\begin{equation}
\textsc{SolveScale} \,=\, \frac{\alpha}{\bar{R}_{\text{solve}}}
\qquad\text{and}\qquad
\textsc{OrderScale} \,=\, \frac{1-\alpha}{\bar{R}_{\text{order}}},
\end{equation}
and use $R_{\text{total}}=\textsc{SolveScale}\cdot R_{\text{solve}} + \textsc{OrderScale}\cdot R_{\text{order}}$ throughout training.

This ensures that, at initialization, each component contributes to $R_{\text{total}}$ in the target ratio regardless of absolute~scale. It simplifies analysis across mixtures by keeping the intended weighting explicit and stable over~runs.

In practice, we run a short evaluation pass of the fine-tuned model on the validation split before GRPO, compute the empirical means of each reward term, and set fixed scaling factors once. We then keep these factors constant for the entire post-training phase. Holding the scales fixed aligns initial magnitudes with the chosen mixture and avoids accidental dominance from raw reward-scale differences, making mixture comparisons interpretable. Under this normalization, improvements in either component contribute according to the intended mixture.

\section{Results and Analysis}

We evaluate our models on the held-out test set using puzzle accuracy as the primary metric---the fraction of puzzles fully solved by the model. All results use greedy decoding at inference time. We report metrics for 
\begin{enumerate}
\item the model fine-tuned on random-order sequences, and
\item models post-trained with GRPO starting from the random-order fine-tuned checkpoint under fixed reward mixtures $\alpha \in \{0.75, 0.9, 0.95, 0.99, 1\}$.
\end{enumerate}

% \subsection{Fine-Tuned Baselines}

% The fine-tuned model trained on random solution order achieves a ${0.282}$ cell accuracy on the test set. In contrast, the model fine-tuned on solver-order sequences reaches ${0.520}$, confirming that access to solver ordering acts as a strong inductive bias \citep{NEURIPS2024_67b31ca1}.

\subsection{GRPO with Different Reward Combinations}

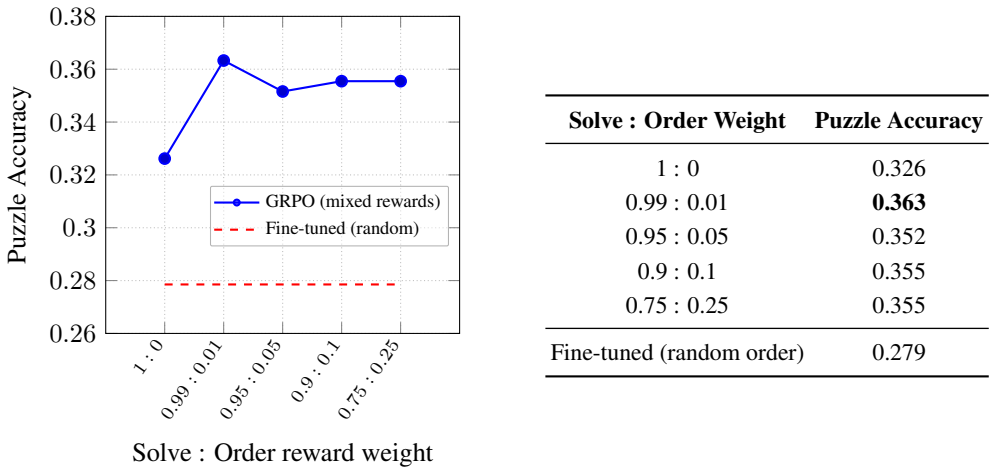
\begin{figure*}[!t]
\centering

% (a) Plot on the left — centered vertically
\begin{subfigure}[c]{0.45\textwidth}
\centering
\begin{tikzpicture}[baseline=(current bounding box.center)]

\begin{axis}[
    width=\linewidth,
    height=5.8cm,
    xlabel={Solve : Order reward weight},
    ylabel={Puzzle Accuracy},
    ymin=0.26, ymax=0.38,
    ytick={0.26,0.28,0.30,0.32,0.34,0.36,0.38},
    yticklabel style={/pgf/number format/fixed,/pgf/number format/precision=2},
    xtick={1,2,3,4,5},
    xticklabels={$1:0$,$0.99:0.01$,$0.95:0.05$,$0.9:0.1$,$0.75:0.25$},
    xticklabel style={font=\scriptsize, rotate=55, anchor=east},
    enlarge x limits=0.25,
    grid=both,
    grid style={densely dotted},
    legend style={
        font=\scriptsize,
        at={(0.97,0.37)},
        anchor=east,
        draw=black!30,
        fill=white,
        fill opacity=0.50,
        text opacity=1,
        rounded corners=0.6pt,
        inner sep=1.0pt
    },
    legend cell align={left},
    legend image post style={mark options={scale=0.55}},
]
\addplot+[mark=*, thick] coordinates {
    (1,0.326171875)
    (2,0.36328125)
    (3,0.3515625)
    (4,0.35546875)
    (5,0.35546875)
};
\addlegendentry{GRPO (mixed rewards)}

\addplot[dashed, thick, red] coordinates { (1,0.2785373) (5,0.2785373) };
\addlegendentry{Fine-tuned (random)}
\end{axis}

\end{tikzpicture}
\label{fig:mix-curve}
\end{subfigure}
\hspace{0.06\textwidth}
%
% (b) Table on the right — centered vertically, larger font kept
\begin{subfigure}[c]{0.40\textwidth}
\centering
\small
\setlength{\tabcolsep}{2pt}
\renewcommand{\arraystretch}{1.30}
\begin{tabular}{cc}
\toprule
\textbf{Solve : Order Weight} & \textbf{Puzzle Accuracy} \\
\midrule
1 : 0          & 0.326 \\
0.99 : 0.01    & \textbf{0.363} \\
0.95 : 0.05    & 0.352 \\
0.9 : 0.1      & 0.355 \\
0.75 : 0.25    & 0.355 \\
\midrule
Fine-tuned (random order) & 0.279 \\
\bottomrule
\end{tabular}
\label{table:rewardmix_table}
\end{subfigure}

\caption{\textbf{Reward mixtures and performance.} Effect of reward mixing on Zebra puzzle accuracy. Each point is GRPO post-trained on the fine-tuned (random order) model at the indicated $\alpha$. (Note: x-axis positions are categorical and not equidistant in $\alpha$.)}
\label{fig:mix-side-by-side}
\end{figure*}

We first fine-tune a model on random solution order, which achieves a ${0.279}$ puzzle accuracy on the test set. We then post-train this model with GRPO using different reward mixtures. Figure~\ref{fig:mix-side-by-side} summarizes test puzzle accuracies. Mixtures that include a non-zero ordering component consistently outperform task-only optimization ($1:0$), with the best result at a ${0.99:0.01}$ solve-to-order weighting (${0.363}$). Notably, even a very small ordering share yields a clear gain over task-only (${0.326}$), suggesting that the ordering signal is effective as a light shaping term. A broader range of mixtures also improves over the fine-tuned baseline (${0.279}$): ${0.75:0.25}$ and ${0.9:0.1}$ both reach ${0.355}$, while ${0.95:0.05}$ reaches ${0.352}$.

Overall, these results on Zebra puzzles suggest that a coarse scalar ordering hint complements the solved objective: mixed rewards steer the model toward solver-like trajectories at inference time without requiring solver-order sequences during fine-tuning.

\subsection{Reward Scaling Effects}

The bootstrapped reward scaling makes reward mixing well-defined by aligning component magnitudes to the chosen mixture at initialization. Without normalization, differences in raw scale can cause $R_{\text{total}}$ to be dominated by one term regardless of $\alpha$, making mixture comparisons misleading. By setting fixed scales from validation-set means, we can mix solved and ordering rewards in a controlled ratio and attribute performance changes to the intended weighting.

\section{Conclusion}

We asked whether a scalar hint about temporal structure (alignment with a canonical solver order) can improve RL post-training without changing supervised data or model architecture. On Zebra puzzles, GRPO with bootstrapped mixtures of solved and ordering rewards yields consistent gains over task-only optimization; even a very small ordering weight improves puzzle accuracy, with the best mixture at $0.99:0.01$.

The ordering reward nudges the policy toward a solver-like progression, producing more canonical rollouts even when the fine-tuning data is randomly ordered. Practically, this provides a cheap, modular post-training knob for injecting structural bias without curating new supervised datasets or training from scratch.

\paragraph{Limitations.}
This work is ongoing. Our experiments are limited to a single task (Zebra puzzles) and a single model (a GPT-2 style Transformer). We also use fixed bootstrapped scaling factors; because reward components can improve at different rates during training, these fixed scales may become miscalibrated as the policy shifts. A natural next step is to test whether periodically updating the scaling factors improves stability and performance, and whether these findings generalize across additional tasks, scales, and architectures.

\clearpage

\bibliography{iclr2026_conference}
\bibliographystyle{iclr2026_conference}

\clearpage

\appendix
\section{Hyperparameters}
\label{sec:appx_hyperparameters}
We list the hyperparameters used to train each setup in this section.

\subsection{Standard Fine-Tuning}
We train a 4-layer model with 4 attention heads per layer, and a hidden size of 256. Table ~\ref{table:hyperparams} lists the training hyperparameters.

\begin{table}[t]
\begin{center}
\begin{tabular}{l|l}
\toprule
\multicolumn{1}{c}{\bf Parameter}  &\multicolumn{1}{c}{\bf Value} \\
\midrule
Learning Rate &   1e-4    \\
Weight Decay  & 0.01 \\
Per-device train batch size & 32 \\ 
Per-device eval batch size & 32 \\
Validation Size & 512 \\
Validation Patience & 10 \\
Optimizer & AdamW \\
\bottomrule
\end{tabular}
\end{center}
\caption{\textbf{Hyperparameters for fine-tuning.} 
}\label{table:hyperparams}
\end{table}

\subsection{GRPO}
Table ~\ref{table:hyperparams_grpo} lists the hyperparameters used for GRPO post-training.

\section{LLM Usage}
We used an LLM (ChatGPT) to assist with writing edits and \LaTeX{} formatting. All technical content, experiments, and claims were produced and verified by the authors.

\begin{table}[t]
\begin{center}
\begin{tabular}{l|l}
\toprule
\multicolumn{1}{c}{\bf Parameter}  &\multicolumn{1}{c}{\bf Value} \\
\midrule
Learning Rate &   1e-6    \\
Per-device train batch size & 128 \\ 
Per-device eval batch size & 128 \\
Number of rollouts & 8 \\
Weight Decay  & 0.01 \\
KL penalty Beta & 0.01 \\
Optimizer & AdamW \\

\bottomrule
\end{tabular}
\end{center}
\caption{\textbf{Hyperparameters for GRPO.} 
}\label{table:hyperparams_grpo}
\end{table}

\end{document}